\documentclass{article}

\pdfoutput=1

\usepackage[preprint, nonatbib]{neurips_2024}

\usepackage[utf8]{inputenc} 
\usepackage[T1]{fontenc}    
\usepackage{hyperref}       
\usepackage{url}            
\usepackage{booktabs}       
\usepackage{amsfonts}       
\usepackage{nicefrac}       
\usepackage{microtype}      
\usepackage{xcolor}         

\usepackage{multirow}
\usepackage{multicol}
\usepackage{makecell}
\usepackage{hhline}
\usepackage{graphicx}
\usepackage{xspace}
\usepackage{subfig}
\usepackage{wrapfig}
\usepackage{tablefootnote}
\usepackage{enumitem}
\usepackage{amsmath}

\newcommand{\imagenet}{\textsc{ImageNet}\xspace}
\newcommand{\lsun}{\textsc{LSUN-Bedroom}\xspace}
\newcommand{\rigid}{\textsc{RIGID}\xspace}

\usepackage{color}

\title{RIGID: A Training-Free and Model-Agnostic Framework for Robust AI-Generated Image Detection}

\author{%
  Zhiyuan~He \\
  Department of Computer Science and Engineering\\
  The Chinese University of Hong Kong\\
  \texttt{zyhe@cse.cuhk.edu.hk} \\
  \And
  Pin-Yu Chen \\
  IBM Research \\
  \texttt{pin-yu.chen@ibm.com} \\
  \AND
  Tsung-Yi Ho \\
  Department of Computer Science and Engineering\\
  The Chinese University of Hong Kong\\
  \texttt{tyho@cse.cuhk.edu.hk} \\
}

\begin{document}

\maketitle

\begin{abstract}

The rapid advances in generative AI models have empowered the creation of highly realistic images with arbitrary content, raising concerns about potential misuse and harm, such as Deepfakes. Current research focuses on training detectors using large datasets of generated images. However, these training-based solutions are often computationally expensive and show limited generalization to unseen generated images.
In this paper, we propose a \textit{training-free} method to distinguish between real and AI-generated images. We first observe that real images are more robust to tiny noise perturbations than AI-generated images in the representation space of vision foundation models. 
Based on this observation, we propose RIGID, a training-free and model-agnostic method for \underline{r}obust A\underline{I}-\underline{g}enerated \underline{i}mage \underline{d}etection. RIGID is a simple yet effective approach that identifies whether an image is AI-generated by comparing the representation similarity between the original and the noise-perturbed counterpart.
Our evaluation on a diverse set of AI-generated images and benchmarks shows that RIGID significantly outperforms existing training-based and training-free detectors. In particular, the average performance of RIGID exceeds the current best training-free method by more than 25\%.
Importantly, RIGID exhibits strong generalization across different image generation methods and robustness to image corruptions.
\end{abstract}

\section{Introduction}
\label{sec:intro}

In recent years, deep learning has revolutionized image generation, enabling the creation of highly realistic images. Platforms such as Stable Diffusion~\cite{sd} and Midjourney~\cite{midjourney}~allow users to generate arbitrary content through text prompts. However,  these advanced Generative AI (GenAI) applications are accomplished with amplified risks and concerns about misuse, such as Deepfakes. Some prompt-based jailbreak techniques~\cite{chin2023prompting4debugging,jailbreak1, jailbreak2} can bypass platforms' safeguards and generate inappropriate content, highlighting the urgent quest for practical solutions to reliable AI-generated image detection.

In the space of AI-generated image detection, a common practice is to design a detector that learns to distinguish between real and generated images. 
Early research~\cite{artifact1, artifact2, artifact3} discovered that the upsampling process in  Generative Adversarial Network (GAN~\cite{gan}) leaves periodic artifacts in the spatial or frequency domain of the generated images, allowing for effective detection of low-quality generated images by checking these specific traces.
However, synthetic artifacts have been weakened with advances in generation methods~\cite{corvi}. This has led to the development of numerous training-based detection methods, which learn common features of generated images by training on large datasets of real and fake images. 
Wang et al.~\cite{wang} show that a deep neural network (DNN) classifier trained on images from a single GAN can surprisingly generalize to images from unseen GANs. 
Gragnaniello et al.~\cite{gragnaniello} enhance detection performance by using extensive data augmentations.
Corvi et al.~\cite{corvi} train a classifier on images generated by Latent Diffusion Model (LDM~\cite{ldm}). Ojha et al.~\cite{ojha} train a simple linear classifier on features extracted from the pretrained CLIP~\cite{clip} model. DIRE~\cite{dire}, on the other hand, computes the diffusion inverse reconstruction error for both real and fake images and trains a detector to distinguish between these errors.

\begin{figure*}[t]
    \includegraphics[width=\linewidth]{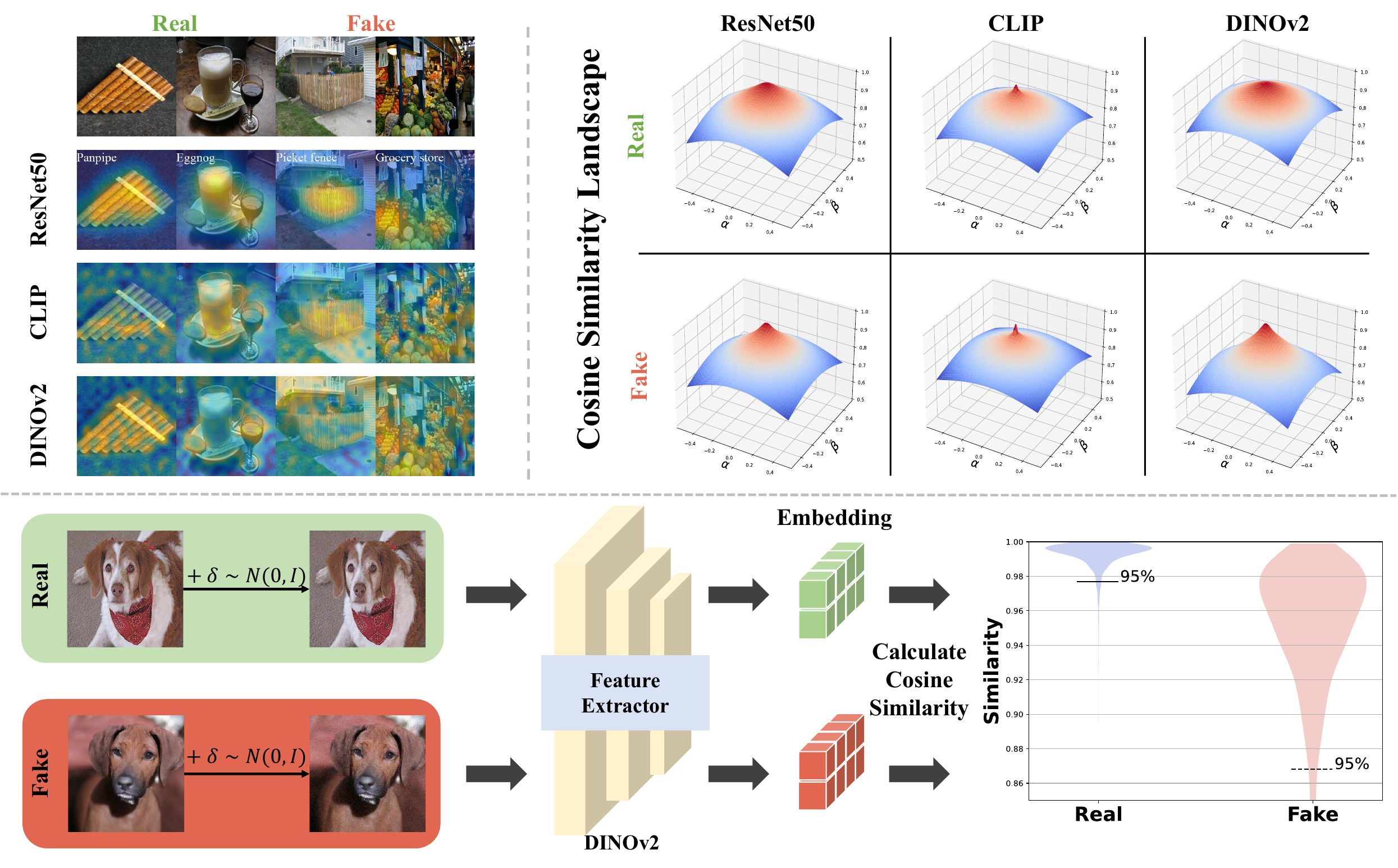}
    \caption{Overview of \textbf{\rigid.} \textbf{Upper left:} visualization of the attention range of different models for real images and AI-generated (fake) images by GradCAM~\cite{gradcam}. CLIP and DINOV2 attend better to global context than ResNet 50. \textbf{Upper right:} visualization of the cosine similarity landscape for real and AI-generated images by plotting the interpolation of two random directions in the image pixel space with coefficients $\alpha$ and $\beta$, following~\cite{landscape}. We find that on DINOv2, real and AI-generated images exhibit distinct sensitivity results. See details of how to plot the landscape in Appendix~\ref{ap:landscape}. \textbf{Bottom:} the framework of \rigid. \rigid uses a pretrained feature extractor to compute the pairwise cosine similarity on the original and noise-perturbed images for AI-generated image detection. The entire detection process is training-free, model-agnostic, and efficient.  See Sec.~\ref{sec:rigid} for details.}
\label{fig:framework}
\vspace{-6mm}
\end{figure*}

While current training-based detectors demonstrate promising results, they still have several limitations. First, their performance is heavily reliant on the quantity, quality, and diversity of the training data. Second, 
the training and re-training costs can be significant and scale unfavorably with the data volume. Finally, the observed drop in their generalization ability to images generated by new or unforeseen models. To circumvent these drawbacks, AEROBLADE~\cite{aeroblade} presents a training-free solution by computing the reconstruction error of a pretrained autoencoder only in the inference phase. Although AEROBLADE only shows good detection performance on images generated by LDM, it opens up new avenues for research in training-free AI-generated image detection.  

In this paper, we aim to develop a more efficient training-free and model-agnostic AI-generated image detection framework. We start by summarizing the lessons from existing studies as a unified paradigm: \textit{the exploration of effective representations contrasting real v.s. AI-generated images is essential to successful detection.}
This exploration has spanned various domains, including the frequency domain of images, the feature space of common classifiers, the representation space of pretrained large vision models, and the reconstruction error space.
However, a crucial question remains: \textit{\textbf{What kind of representation space is most suitable for detecting AI-generated images?}} 

Stein et al.~\cite{exposing} argue that models that consider both global image structure and key objective allow for a richer evaluation of a generative model. 
Motivated by this observation, we visualize the heatmap of different vision models by GradCAM~\cite{gradcam} on some images (upper left of Fig.~\ref{fig:framework}). 
The results demonstrate that supervised models (ResNet 50~\cite{resnet}) focus primarily on the main objects directly relevant to the classification result. In contrast, self-supervised models, particularly DINOv2~\cite{dinov2}, exhibit a more holistic perspective, capturing a broader understanding of the image content \cite{paul2022vision}.
Furthermore, we investigate the sensitivity of real and fake images to small perturbations, with a plot of the cosine similarity landscape (see Sec. \ref{sec:rigid} for details) shown in the upper right of Fig.~\ref{fig:framework}.
Our findings reveal that, compared to real images, AI-generated images exhibit higher sensitivity to small perturbations when using models like DINOv2, which adopts a more global view. Interestingly, this phenomenon is not so obvious in ResNet 50 and CLIP. The reason could be that DINOv2 uses self-supervised learning on images only, while ResNet 50 uses image labels for supervised learning, and CLIP uses image captions for weakly supervised learning.

Taking advantage of this unique sensitivity property, we propose a \textbf{\underline{R}obust A\underline{I}-\underline{G}enerated \underline{I}mage \underline{D}etection} method, \textbf{\rigid}. 
RIGID is a simple and efficient detection method. As shown in the bottom of Fig.~\ref{fig:framework}, given an image, RIGID can effectively tell if it is real or AI-generated, 
by only adding some minor noise and calculating the cosine similarity between the original and the noisy images to set a detection threshold. 
Notably, RIGID \textbf{does not require any training or a priori knowledge of the generated images (e.g., which model is used for generation)}. 
We evaluate the detection performance of RIGID on a wide range of AI-generated image datasets and benchmarks. The results show that RIGID, albeit a training-free method, is often more effective than extensively trained classifiers. 
Moreover, RIGID outperforms the state-of-the-art (SOTA) training-free method AEROBLADE by more than \textbf{25\%} in terms of average precision.
Furthermore, RIGID exhibits strong generalization across various generative methods and robustness to common image corruptions.

We summarize our \textbf{main contributions} as follows:
\begin{itemize}[leftmargin=*]
    \item We propose RIGID, a simple training-free method for detecting AI-generated images.
    \item We prove that the detection mechanism in RIGID is equivalent to comparing the gradient norm (i.e., sensitivity) of a smoothed cosine similarity metric, as illustrated by Fig.~\ref{fig:framework} (top right panel).  
    \item Experiments show that RIGID outperforms the SOTA training-free method, is mostly more effective than training-based detectors, and has strong generalization across image generation models and robustness to image corruptions.

\end{itemize}

\section{Related Works}
\label{sec:related}

\textbf{Image Generation.}
GANs and diffusion models are mainstream techniques for image generation. Among them, BigGAN~\cite{biggan} applies orthogonal regularization to the generator to improve training stability, and StyleGAN~\cite{stylegan} further improves the controllability of generated images by incorporating a style-based generator. Models such as the Denoising Diffusion Probabilistic Model (DDPM~\cite{ddpm}) and LDM~\cite{ldm} have shown impressive results in generating high-quality images.
Another line of research focuses on conditional image generation, which refers to generating images based on specific input conditions (such as text descriptions or semantic labels). GigaGAN~\cite{gigagan} combines CLIP~\cite{clip} and GAN to achieve text-to-image generation.
The ablative diffusion model (ADM~\cite{adm}) achieves an efficient text-to-image generation architecture by removing the self-attention mechanism.
Diffusion-based Transformer (DiT~\cite{dit}) replaces U-Net in LDM with Transformer and uses Transformer's ability to capture global context to improve the quality of text-to-image generation.
These methods give rise to popular text-to-image generation tools such as Stable Diffusion~\cite{sd} and Midjourney~\cite{midjourney}.

\textbf{AI-generated Image Detection.}
Early efforts focused on leveraging hand-crafted features, such as color cues~\cite{color}, saturation cues~\cite{saturation}, and co-occurrence features~\cite{cooccurrence}, to identify machine-edited images. However, these features are no longer reliable indicators, as modern generative models have largely overcome these limitations. Another successful strategy is to analyze images in the frequency domain~\cite{artifact1, artifact2, artifact3}, where the generated images exhibit distinguishable artifacts. However, these artifacts are only evident in the upsampling model and cannot be used to detect images generated by diffusion models~\cite{corvi}. 
Recently, various learning-based approaches have been proposed.
Wang et al.~\cite{wang} demonstrated that a simple classifier trained on ProGAN-generated~\cite{progan} images, augmented with Gaussian blurring and JPEG compression, could generalize to other unseen GAN-generated images. Gragnaniello et al.~\cite{gragnaniello} further improved detection performance by employing more extensive data augmentations. Corvi et al.~\cite{corvi} extended Wang's approach to diffusion models. Ojha et al.~\cite{ojha} explored leveraging pretrained CLIP features to train a linear classifier. DIRE~\cite{dire} finds that diffusion models can reconstruct diffusion-generated images more accurately than real images, utilizing the reconstruction error to train the detector. 
However, these training-based methods generally suffer from limited generalization and require computationally expensive training processes. 
This has led to a growing interest in training-free detection methods.
AEROBLADE~\cite{aeroblade} detects generated images solely based on the reconstruction error of the image passing through an autoencoder. Nevertheless, it is only effective for images generated by LDM using similar autoencoders, and its generalizability remains a challenge.

\section{Methodology}
\label{sec:method}

\subsection{\rigid}
\label{sec:rigid}

\textbf{Design Objective.}
This work aims to develop an effective training-free method for detecting AI-generated images. Unlike existing training-free methods like AEROBLADE~\cite{aeroblade}, which rely on the autoencoder used by LDM, our goal is to achieve effective detection across images produced by various generative methods without any prior knowledge of the generation process (i.e., a model-agnostic detector). 
Notably, our approach does not change any component of the pretrained model, including the architecture and training weights. Its detection solely uses the inference results of an off-the-shelf pretrained feature extractor to derive features differentiating real and generated images.

\textbf{Core Idea.} 
While real and generated images often exhibit subtle differences in semantics and texture, these distinctions become increasingly difficult to discern by a human user as generation methods advance. Current training-based detectors attempt to extract these hidden differences through supervised learning. Our work takes a different approach by exploiting the sensitivity difference of real and generated images to small perturbations. 
As shown in the upper right of Fig.~\ref{fig:framework}, adding noise perturbations causes the features of real images to change continuously, resulting in a smoother gradient.
Conversely, generated images are more sensitive to noise,  leading to a steeper change and gradient.
Although the added noise is subtle, it can act as a probe for global features covering texture-rich and texture-poor regions of the image, which proves 
 beneficial for generated image detection~\cite{patch}.
To accurately perceive how global features are affected by noise, 
we employ DINOv2~\cite{dinov2} as our backbone model (feature extractor) since it has a holistic image view~\cite{exposing}. A detailed discussion on the impact of different backbones on detection performance is provided in Sec.~\ref{sec:ablation}.

\textbf{Workflow.} 
The workflow of RIGID is illustrated at the bottom of Fig.~\ref{fig:framework}. Our proposed AI-generated image detector leverages the sensitivity difference between real and fake images to tiny perturbations for classification. Given an input sample, RIGID begins by adding subtal perturbations to the image. Then, both the original input sample and its noise-perturbed counterpart are fed into DINOv2 to obtain their feature embeddings. Next, the cosine similarity of the embedding is calculated and used to determine whether the input is a generated image through the following threshold-based detection:
\begin{equation}
    S(x) = \mathbf{1}\{\textsf{sim}(f(x), f(x+ \lambda \cdot \delta)) \leq \epsilon \}; \qquad \delta \sim N(0,I)
    \label{eq:rigid}
\end{equation}
where $f(\cdot)$ is the feature extractor, $\textsf{sim}(\cdot)$ represents the cosine similarity between two embeddings, $\mathbf{1}\{\cdot \}$ denotes the binary indicator function, $\delta$ is the additive noise drawn from a standard normal distribution $N(0,I)$, and $\lambda$ controls the noise level. An image is classified as AI-generated when the cosine similarity between the embeddings of the input image and its noised counterpart falls below a specified threshold $\epsilon$. The threshold $\epsilon$ is typically chosen to ensure the correct classification of the majority of real images (e.g., 95\%). Notably, the selection of these thresholds is independent of the generated images. Compared to existing methods, our approach offers several significant advantages:
\begin{itemize}[leftmargin=*]
    \item \textbf{Training-free:} RIGID operates solely during the inference phase, eliminating the expensive training costs like~\cite{corvi, wang, gragnaniello, dire}.
    \item \textbf{Generation-Independent:} Unlike AEROBLADE~\cite{aeroblade}, a training-free method that relies on an autoencoder closely tied to the underlying image generation model, RIGID utilizes DINOv2~\cite{dinov2}, a model trained with self-supervised learning without generated images.
    \item \textbf{Model-agnostic:} RIGID does not assume the knowledge of image generation models, demonstrating the capability to detect a wide range of AI-generated images.
    \item \textbf{Computationally Efficient:} Unlike DIRE~\cite{dire} and AEROBLADE~\cite{aeroblade}, which need to compute reconstruction errors involving multi-step forward and backward diffusion processes via diffusion models, RIGID operates more efficiently by calculating embedding similarity directly.
\end{itemize}

\subsection{Theorectical Analysis}
Based on our RIGID framework, given a backbone $f(\cdot): \mathbb{R}^n \rightarrow \mathbb{R}^d$ and the cosine similarity function $h(\cdot): \mathbb{R}^{d} \times \mathbb{R}^{d } \rightarrow \mathbb{R}$. The score function in eq.~\ref{eq:rigid} can be reformulated in expectation as: 
\begin{equation}
    \centering
    G(x) = ((h \circ f)*N(0, \lambda^2 I))(x) = \mathbb{E}_{\delta \sim N(0, \lambda^2 I)}[h(f(x+\delta), f(x))]
\label{eq:theory1}
\end{equation}
where $*$ denotes the convolution operator between two functions, defined as $h*g = \int_{\mathbb{R}^d} h(t)g(x-t)dt$.
Then, according to the Stein’s lemma~\cite{stein}, $G(x)$ is differentiable with a gradient of:
\begin{equation}
\begin{aligned}
    \nabla G(x) &= \frac{1}{\left(2 \pi \lambda^2\right)^{d / 2}} \int_{\mathbb{R}^d} (h \circ f)(t) \frac{t-x}{\lambda^2} \exp \left(\frac{1}{2 \lambda^2}\|x-t\|_2^2\right) d t \\
    &= \frac{1}{\lambda^2} \mathbb{E}_{\delta \sim \mathcal{N}\left(0, \lambda^2 I\right)}[\delta \cdot h(f(x+\delta), f(x))]
\end{aligned}
\end{equation}
Therefore, the random perturbation $\delta$ introduced by RIGID to $f(x+\delta)$ can be viewed as an operation of probing the gradient of the smoothed cosine similarity metric $G(x)$.
According to the cosine similarity landscape in the upper right panel of Fig.~\ref{fig:framework}, the gradient norm of fake images is greater than that of real images due to higher sensitivity to random perturbations. This analysis shows that  \rigid is effectively leveraging the gradient information of the cosine similarity metric for detection.

\section{Experiments}
\label{sec:experiment}

\subsection{Setup}


\textbf{Dataset.} To provide a comprehensive evaluation of AI-generated image detectors, we deviated from previous studies that often limited their testing to a single dataset or generation method. We designed two rigorous test sets to assess the performance of these detectors across a diverse range of generative models and datasets.
First, following the work of~\cite{exposing}, we evaluate the detectors' performance on two widely used datasets: \imagenet~\cite{imagenet} and \lsun~\cite{lsun}. We selected a variety of generative methods representing different model architectures, including Diffusion Models, GANs, variational autoencoders (VAEs), Transformer-based models, and Mask Prediction models. These methods are chosen from a leaderboard of generated images~\cite{leaderboard}, ensuring the representation of SOTA generative capabilities.
Specifically, on \imagenet, we choose ADM~\cite{adm}, ADM-G, LDM~\cite{ldm}, DiT-XL2~\cite{dit}, BigGAN~\cite{biggan}, GigaGAN~\cite{gigagan}, StyleGAN~\cite{stylegan}, RQ-Transformer~\cite{rqtransformer}, and Mask-GIT~\cite{maskgit}. For \lsun, we select ADM, DDPM~\cite{ddpm}, iDDPM~\cite{iddpm}, Diffusion Projected GAN~\cite{projdiffusiongan}, Projected GAN~\cite{projdiffusiongan}, StyleGAN~\cite{stylegan}, and Unleasing Transformer~\cite{unleashing}. 
Each model generated 100k images, with the same number of images per class for class-conditional models. 
In addition, we expand our evaluation to images generated by popular generative platforms, including Stable Diffusion 1.4 and 1.5~\cite{sd}, Midjourney~\cite{midjourney}, and Wukong~\cite{wukong}. These images are collected from GenImage~\cite{genimage}, a recently established benchmark for AI-generated image detection. A detailed description of the datasets used in our evaluation can be found in Appendix~\ref{ap:datasets}.

\textbf{Evaluation Metrics.} Following existing detection methods~\cite{corvi, wang}, we primarily utilize two key metrics to evaluate the performance of the detectors in our experiments: Area Under the Receiver Operating Characteristic curve (AUC) and Average Precision (AP). Both AUC and AP provide a quantitative measure of detection accuracy, with higher scores indicating better performance.

\textbf{Baselines.}
We conducted a comparative analysis of \rigid against a range of established AI-generated image detection methods, encompassing both training-based and training-free approaches. The former include Wang et al~\cite{wang}, Gragnaniello et al~\cite{gragnaniello}, Corvi et al~\cite{corvi}, and DIRE~\cite{dire}. 
The latter includes a prominent training-free method: AEROBLADE~\cite{aeroblade}. Detailed information regarding the implementation of these baseline methods can be found in Appendix~\ref{ap:baseline}.

\subsection{Evaluation of Detection Performance}

\subsubsection{Comparison with Baselines}

We conducted a comprehensive comparative analysis of various AI-generated image detection methods,  evaluating their performance on \imagenet and \lsun, as presented in Table~\ref{tb:imagenet} and \ref{tb:lsun}, respectively. Our analysis revealed several key findings of \rigid:

\textbf{Superior Performance.}
\rigid consistently demonstrated exceptional performance across both datasets. Notably, it significantly outperformed AEROBLADE, another training-free method, by an average of over 25\%, establishing a new SOTA for training-free detection. Furthermore, \rigid generally surpassed the performance of training-based methods, only falling slightly short for a few specific generative methods.

\begin{table*}[t]
\small
\caption{The AUC and AP of different AI-generated image detectors on \imagenet. A higher value indicates better performance.
The \textbf{bolded} values are the best performance, and the \underline{\textit{underlined italicized}} values are the second-best performance. The same annotation holds for all tables.}
\setlength\tabcolsep{2pt}
\renewcommand{\arraystretch}{1.3}
\resizebox{1.\linewidth}{!}{
\begin{tabular}{lccccccccccc}
\hline
\multirow{2}{*}{AUC/AP   (\%)} & \multirow{2}{*}{\begin{tabular}[c]{@{}c@{}}Training\\ Samples\end{tabular}} & \multicolumn{4}{c}{Diffusion}                         & \multicolumn{3}{c}{GAN}                 & \multicolumn{2}{c}{VAE}      & \multirow{2}{*}{Average} \\
                               &                                                                             & ADM         & ADMG        & LDM         & DiT         & BigGAN      & GigaGAN     & StyleGAN XL & RQ-Transformer & Mask GIT    &                       \\ \hline 
Wang                           & 720 000                                                                     & \underline{\textit{65.96}}/\underline{\textit{66.75}} & \underline{\textit{65.56}}/\underline{\textit{66.59}} & \underline{\textit{67.82}}/\underline{\textit{69.43}} & 61.97/64.25 & \underline{\textit{83.15}}/\underline{\textit{84.76}} & \underline{\textit{71.19}}/\underline{\textit{69.96}} & \underline{\textit{66.63}}/\underline{\textit{66.06}} & 60.66/61.67    & \underline{\textit{65.43}}/\underline{\textit{66.97}} & \underline{\textit{67.60}}/\underline{\textit{68.43}}           \\
Gragnaniello                   & 400 000                                                                     & 60.21/59.91 & 59.45/59.71 & 61.61/61.37 & 56.67/56.56 & 59.62/58.49 & 53.63/52.35 & 51.58/52.35 & 56.49/54.34    & 53.70/52.68 & 56.99/56.24           \\
Corvi                          & 400 000                                                                     & 63.94/63.85 & 65.55/65.19 & 62.18/60.83 & 56.64/55.23 & 61.91/59.95 & 50.15/49.18 & 48.48/48.05 & 63.21/60.48    & 61.19/59.51 & 59.25/58.03           \\
DIRE                           & 80 000                                                                      & 57.79/56.67 & 57.09/56.80 & 61.47/62.15 & 53.21/53.52 & 49.63/50.00 & 50.00/51.14 & 52.91/53.87 & 53.17/52.41    & 49.93/51.57 & 53.91/54.24           \\ \hline 
AEROBLADE                      & Training Free                                                                           & 52.20/53.65 & 59.24/57.93 & 62.97/61.96 & \textbf{72.98}/\textbf{73.65} & 50.07/50.94 & 55.21/54.87 & 51.17/52.85 & \underline{\textit{70.23}}/\underline{\textit{69.36}}    & 59.80/58.71 & 59.32/59.33           \\
\rigid                          & Training Free                                                                           & \textbf{87.75}/\textbf{86.06} & \textbf{83.50}/\textbf{81.46} & \textbf{81.50}/\textbf{80.23} & \underline{\textit{72.07}}/\underline{\textit{69.55}} & \textbf{93.86}/\textbf{93.57} & \textbf{89.29}/\textbf{87.92} & \textbf{85.94}/\textbf{84.75} & \textbf{93.39}/\textbf{93.11 }   & \textbf{92.65}/\textbf{91.91} & \textbf{86.67}/\textbf{85.40 }          \\ \hline 
\end{tabular}}
\label{tb:imagenet}
\vspace{-5mm}
\end{table*}

\begin{table}[t]
\small
\caption{The AUC and AP of different AI-generated image detectors on \lsun. }
\setlength\tabcolsep{2pt}
\renewcommand{\arraystretch}{1.3}
\resizebox{1.\linewidth}{!}{
\begin{tabular}{lccccccccc}
\hline
AUC/AP   (\%) & \begin{tabular}[c]{@{}c@{}}Training\\ Samples\end{tabular} & ADM         & DDPM        & iDDPM       & \begin{tabular}[c]{@{}c@{}}Diffusion\\ Projected\\ GAN\end{tabular} & \begin{tabular}[c]{@{}c@{}}Projected\\ GAN\end{tabular} & StyleGAN    & \begin{tabular}[c]{@{}c@{}}Unleashing\\ Transformer\end{tabular} & Average        \\ \hline
Wang          & 720 000                                                    & \underline{\textit{66.13}}/\underline{\textit{65.96}} & \underline{\textit{81.87}}/\underline{\textit{82.07}} & \underline{\textit{78.46}}/\underline{\textit{79.13}} & \underline{\textit{90.63}}/\underline{\textit{90.59}}                                                         & \underline{\textit{92.55}}/\underline{\textit{92.43}}                                             & \textbf{98.47}/\textbf{98.34} & \textbf{92.55}/\textbf{92.66}                                                      & \underline{\textit{85.81}}/\underline{\textit{85.88}} \\
Gragnaniello  & 400 000                                                    & 55.92/57.46 & 65.58/65.99 & 62.47/62.87 & 59.15/57.95                                                         & 63.36/62.36                                             & 67.08/66.01 & 66.12/67.00                                                      & 62.96/62.81 \\
Corvi         & 400 000                                                    & 56.67/58.21 & 68.67/70.02 & 68.70/69.57 & 55.46/54.94                                                         & 54.54/55.16                                             & 54.26/55.71 & 72.44/71.91                                                      & 61.54/62.22 \\
DIRE\tablefootnote{Our implementation of DIRE yielded significantly poorer results than originally reported. This discrepancy arises from a format bias in the original method, where real images are JPEG compressed while generated images are stored as lossless PNGs. This bias inflates DIRE's performance, which is discussed in detail in~\cite{aeroblade}.}          & 80 000                                                     & 56.36/57.26 & 60.29/60.87 & 63.52/63.74 & 56.31/55.89                                                         & 57.42/58.14                                             & 58.38/58.83 & 64.77/65.26                                                      & 59.58/60.00 \\ \hline 
AEROBLADE     & Training Free                                                          & 58.03/59.33 & 73.92/74.31 & 68.20/69.18 & 51.46/50.00                                                         & 52.10/50.81                                             & 52.60/50.81 & 61.19/58.34                                                      & 59.46/58.98 \\
RIGID         &  Training Free                                                        & \textbf{74.04}/\textbf{72.92} & \textbf{89.30}/\textbf{89.76} & \textbf{85.61}/\textbf{86.07} & \textbf{93.86}/\textbf{94.49}                                                         & \textbf{94.41}/\textbf{94.81}                                             & \underline{\textit{84.12}}/\underline{\textit{81.53}} & \underline{\textit{92.49}}/\underline{\textit{92.63}}                                                      & \textbf{87.69}/\textbf{87.47} \\ \hline 
\end{tabular}}
\label{tb:lsun}
\end{table}

\textbf{Strong Generalization Ability.}
RIGID exhibited strong generalization capabilities, effectively detecting images generated by diverse methods on both \imagenet and \lsun. This is a significant advantage over existing methods, particularly training-based approaches.
For instance, Wang et al.'s method, trained on ProGAN-generated images, showed a significant performance drop when tested on diffusion-based models compared to GAN-based models. Similarly, Corvi et al.'s method, trained on LDM-generated images, performed poorly on GigaGAN and StyleGAN, approaching random guessing. This highlights a major limitation of training-based methods: their performance is heavily dependent on the training dataset's size and diversity, a point we will elaborate on in Sec.~\ref{sec:discuss}.

\textbf{Independence from Generation Bias.}
Unlike AEROBLADE, which relies on the autoencoder from the generative model to compute reconstruction loss, RIGID operates independently of the underlying generation model in detection. AEROBLADE's performance is inherently tied to the pretrained autoencoder, which is evident in its improved performance on images generated by methods using autoencoders (LDM, DiT, RQ-Transformer). In contrast, RIGID relies solely on DINOv2, a self-supervised vision transformer, making it entirely independent of the specific generative model.

In summary, our results validate the superior performance and generalization capabilities of \rigid for AI-generated image detection, surpassing existing training-based and training-free methods.

  

\begin{figure}[t]
    \begin{minipage}[c]{0.35\textwidth}
        \centering
        \includegraphics[width=0.95\textwidth]{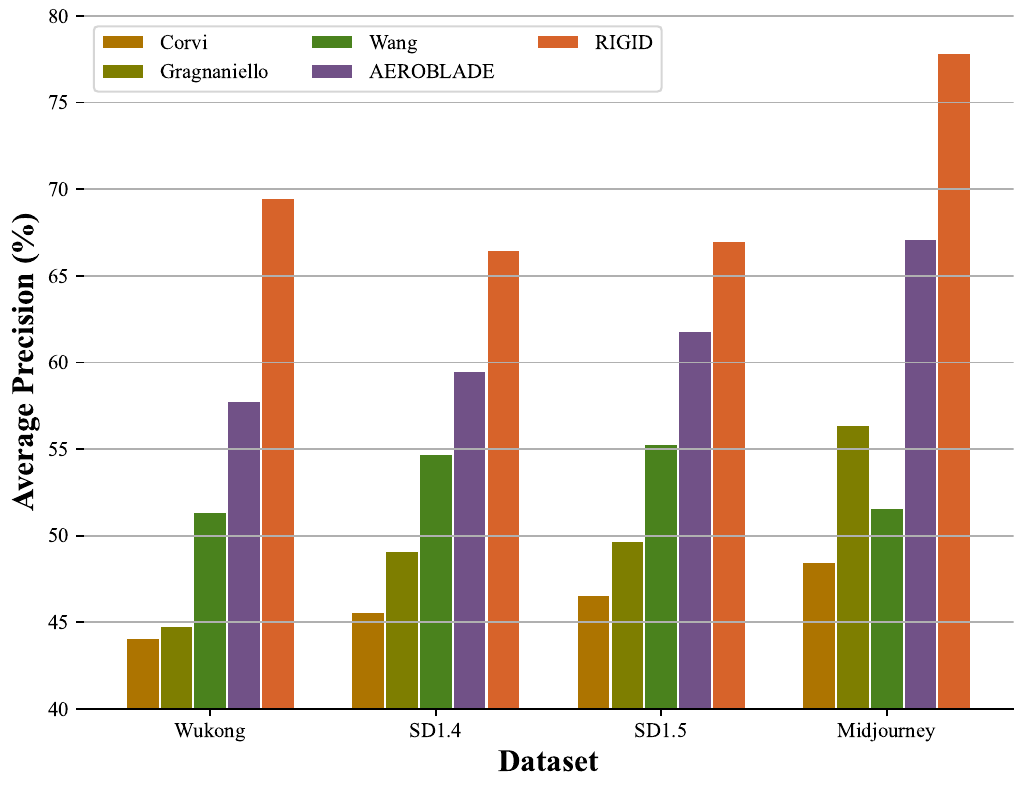}
      \caption{The average precision of various AI-generated image detectors on images generated by popular platforms (Wukong, SD1.4, SD1.5, and Midjourney). }
      \label{fig:mainstream}
    \end{minipage}
    \hspace{1mm}
    \begin{minipage}[c]{0.64\textwidth}
        \vspace{-8mm}
        \centering
        \subfloat[Real: \imagenet; Fake: \lsun]{
    	\includegraphics[width=0.48\textwidth]{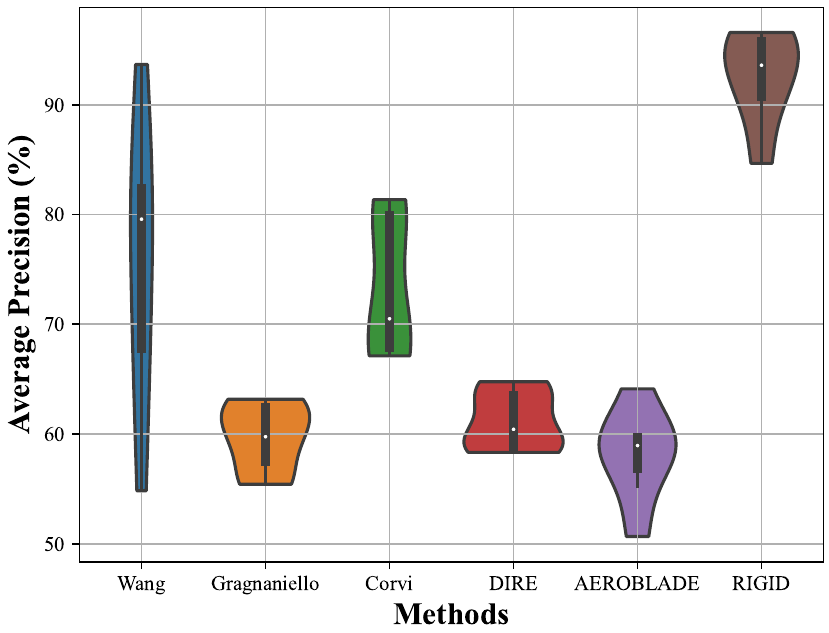}
    	}%
        \subfloat[Real: \lsun; Fake: \imagenet]{
    	\includegraphics[width=0.48\textwidth]{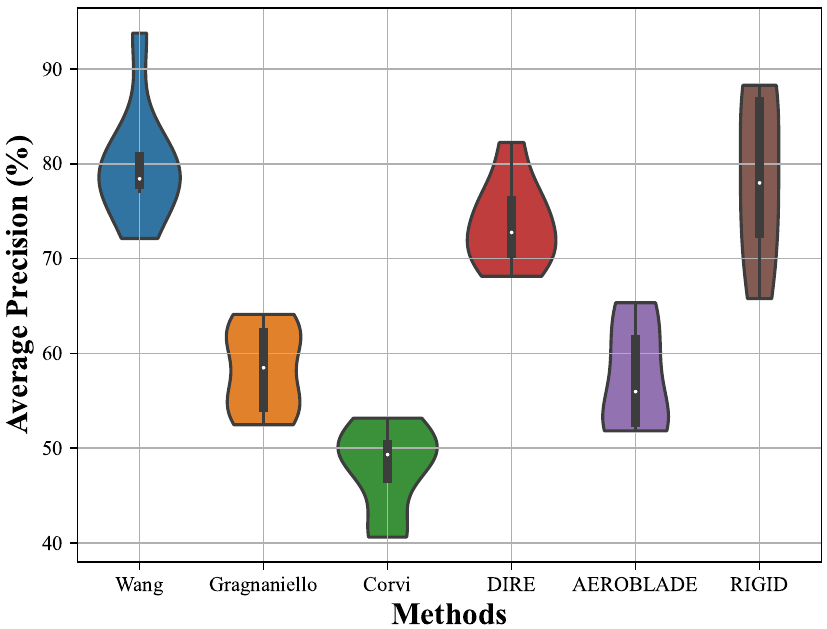}
    	}
        \caption{\textbf{Cross-dataset Evaluation} on \imagenet and \lsun. The violin graph shows AP distribution, where the black bar in the center indicates the interquartile range and the white dot is the median.}
        \label{fig:cross}
        
\end{minipage}
\end{figure}

\subsubsection{Evaluation on Popular Text-to-Image Generation Platforms}

Fig.~\ref{fig:mainstream} compares the detection performance of \rigid and other detection methods on images generated by four widely used platforms: Wukong~\cite{wukong}, SD 1.4~\cite{sd}, SD 1.5 and Midjourney~\cite{midjourney}. All images are extracted from the GenImage benchmark~\cite{genimage}. In this setting, 
we observe that training-free methods outperform training-based methods. This discrepancy arises because the generative models used to synthesize images for training detectors inevitably lag behind the rapidly evolving mainstream generation techniques, which highlights the importance of exploring effective, stable, and training-free detection methods. 
Notably, \rigid consistently outperforms all other methods across four generation platforms, achieving the highest AP scores, with an average performance approximately 10\% higher than AEROBLADE.  
This underscores RIGID's robust performance and generalizability across different types of generated images and models.

\subsubsection{Cross Domain Testing}
Referring to~\cite{dire}, we evaluate the performance of various AI-generated image detection methods under domain shifting, specifically testing scenarios where the training and test data come from different datasets.
Fig.~\ref{fig:cross} presents the results of this evaluation.
In Fig.~\ref{fig:cross} (a), the real images are from \imagenet and the generated (fake) images are from \lsun, while Fig.~\ref{fig:cross} (b) reverses this order. 
Across both scenarios, the performance of \rigid remains remarkably stable even when the training and test data are drawn from different domains, demonstrating its robustness to domain shifts. In contrast, other methods, particularly training-based approaches, exhibit a significant decline in AP when evaluated on a dataset different from their training data. This vulnerability to dataset shift stems from their inherent dependence on the specific characteristics of the training data. Interestingly, both Wang et al.'s method and RIGID show improved performance when real images are sourced from \imagenet and generated images are from \lsun. We attribute this observation to the greater diversity of \imagenet compared to \lsun.

\begin{figure*}[t]
    \includegraphics[width=\linewidth]{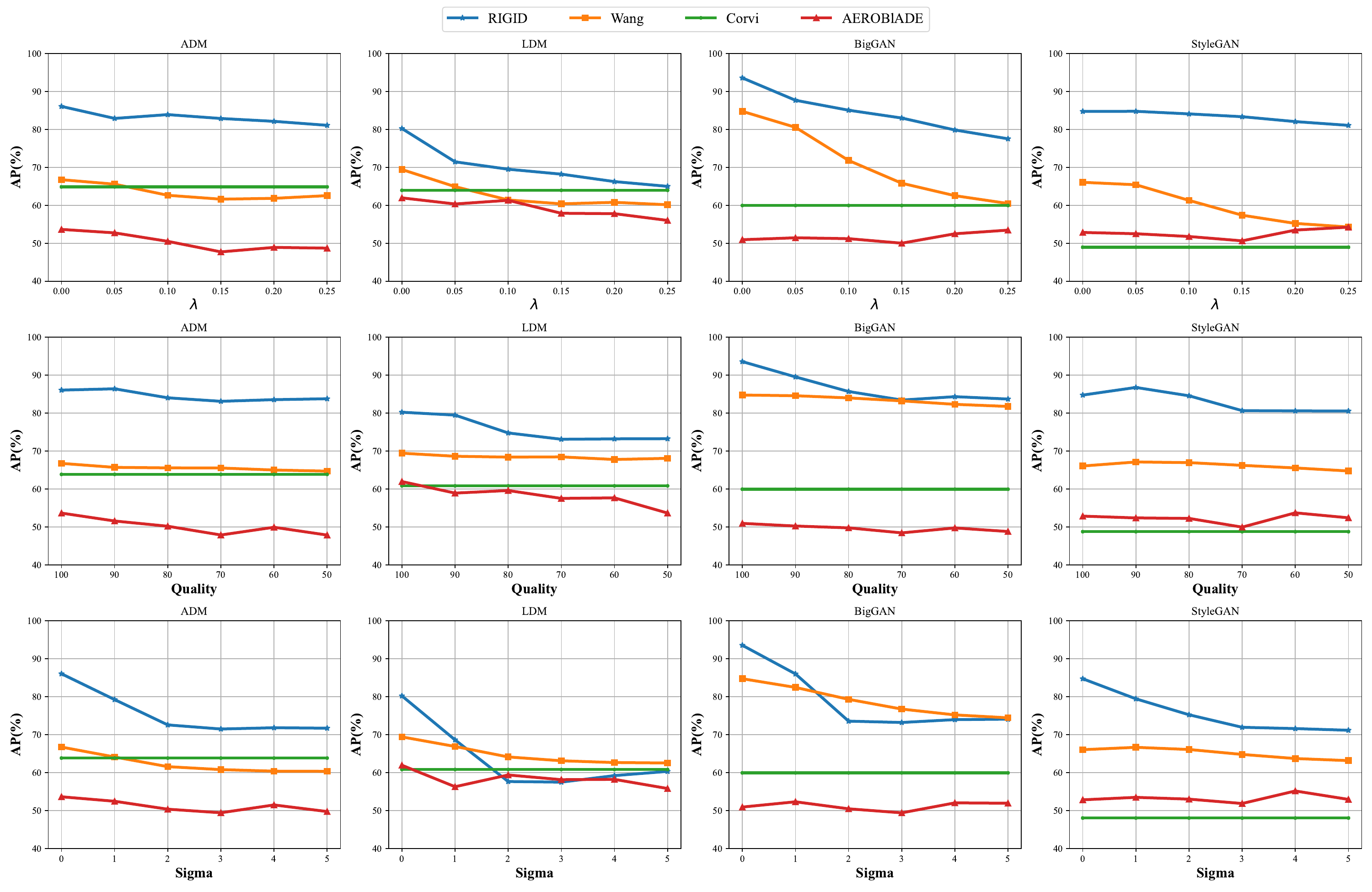}
\caption{\textbf{Robustness to Image Corruptions.} The top row shows the robustness to Gaussian noise ($\lambda$ represents the noise intensity). The second row shows the robustness to JPEG compression, and the bottom row shows the robustness to Gaussian blur. }
\label{fig:robustness}
\end{figure*}

\subsection{Robustness to Image Corruptions}

In real-world scenarios, images are usually subject to various corruptions. Therefore, we follow~\cite{dire, aeroblade} to evaluate the robustness of the detector to three types of image corruptions. As shown in Fig.~\ref{fig:robustness}, each row represents a common image corruption, from top to bottom, Gaussian noise, JPEG compression, and Gaussian blur. We set five levels for each corruption ($\lambda=\{0.05, 0.1, 0.15, 0.2, 0.25\}$; Quality$=\{90, 80, 70, 60, 50\}$; Sigma$=\{1, 2, 3, 4, 5\}$).
The evaluation is performed on four generation methods: ADM~\cite{adm},  LDM~\cite{ldm}, BigGAN~\cite{biggan}, and StyleGAN~\cite{stylegan}. 

We observe that RIGID consistently outperformed baseline methods in most cases, demonstrating greater resilience to these corruptions. In particular, RIGID maintains a significant performance advantage over its training-free counterpart AEROBLADE across all three corruption types for the four generation models.
Notably, training-based methods show less degradation under JPEG compression and Gaussian blur. This can be attributed to the inclusion of these corruptions as augmentations during their training process. However, their performance significantly dropped when faced with unseen corruptions like Gaussian noise. 
For instance, Wang et al.'s method experienced a mere 3\% drop with JPEG compression but a substantial 13\% drop with Gaussian noise.
Therefore, RIGID shows robustness to common image corruptions without training, highlighting the reliability of RIGID and its potential for practical applications where image quality may be compromised.

\subsection{Ablation Studies}
\label{sec:ablation}

\begin{figure}[t]
    \begin{minipage}[c]{0.45\textwidth}
        \centering
        \vspace{5mm}
        \includegraphics[width=\textwidth]{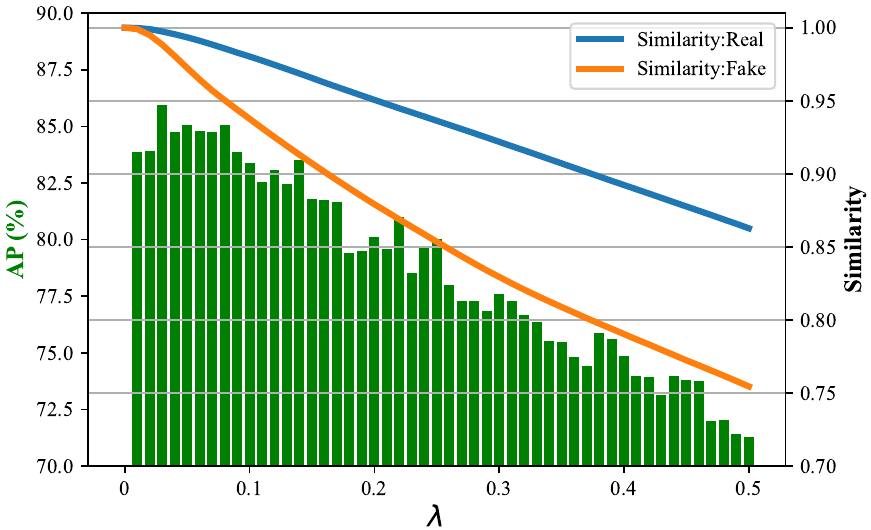}
        \caption{\textbf{Detection performance for different noise intensities (the value $\lambda$ in eq.~\ref{eq:rigid}).} The left/right y-axis is AP/Cosine-Similarity.}
        \label{fig:scaler}
    \end{minipage}
    \hspace{1mm}
    \begin{minipage}[c]{0.54\textwidth}
        \centering
        \includegraphics[width=\textwidth]{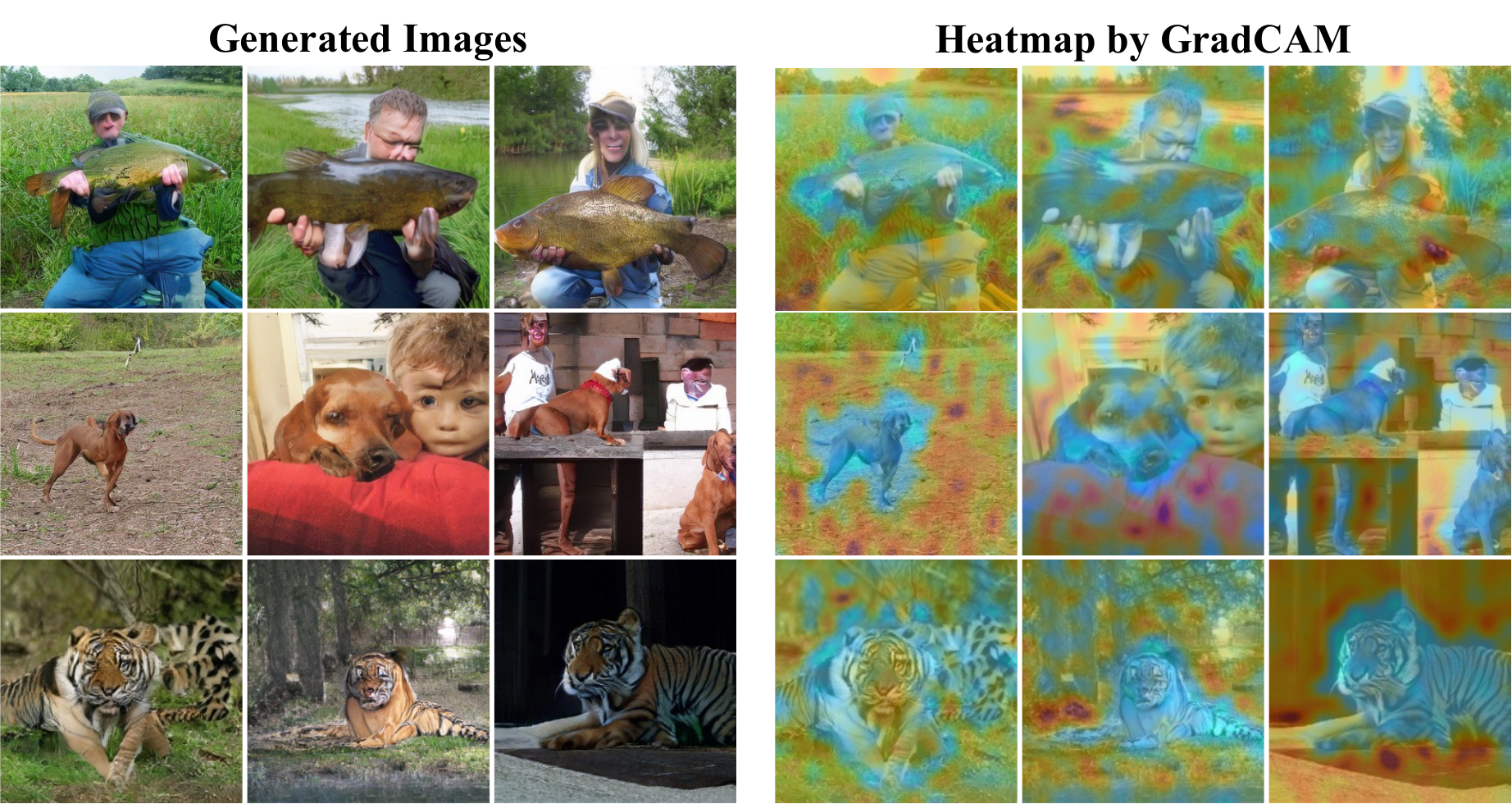}
        \caption{\textbf{Display of AI-generated image attribution.} Note that higher/lower heat levels represent areas identified as real/AI-generated by GradCAM using RIGID.}
        \label{fig:sim_cam}
    \end{minipage}
\end{figure}

\textbf{Noise Intensity.} 
Fig.~\ref{fig:scaler} illustrates the impact of noise intensity ($\lambda$) on RIGID's performance, alongside the trend of cosine similarity between real and generated (fake) images. At $\lambda$ = 0, both real and generated images exhibit a cosine similarity of 1, resulting in an AP of approximately 50\%, equivalent to a random guesser.
As noise intensity increases, the disparity in cosine similarity between real and generated images widens. However, excessively high noise levels negatively impact RIGID's detection performance, likely due to the disruption of normal feature representation caused by the noise. Within a moderate noise range (0 to 0.17), RIGID maintains high detection performance with AP scores greater than or equal to 80\%.
Importantly, even under very high noise levels, RIGID continues to outperform the baseline methods listed in Table~\ref{tb:imagenet}. 
This demonstrates that RIGID is not a hyperparameter-sensitive method.

\begin{figure}[t]
    \centering
    \includegraphics[width=\textwidth]{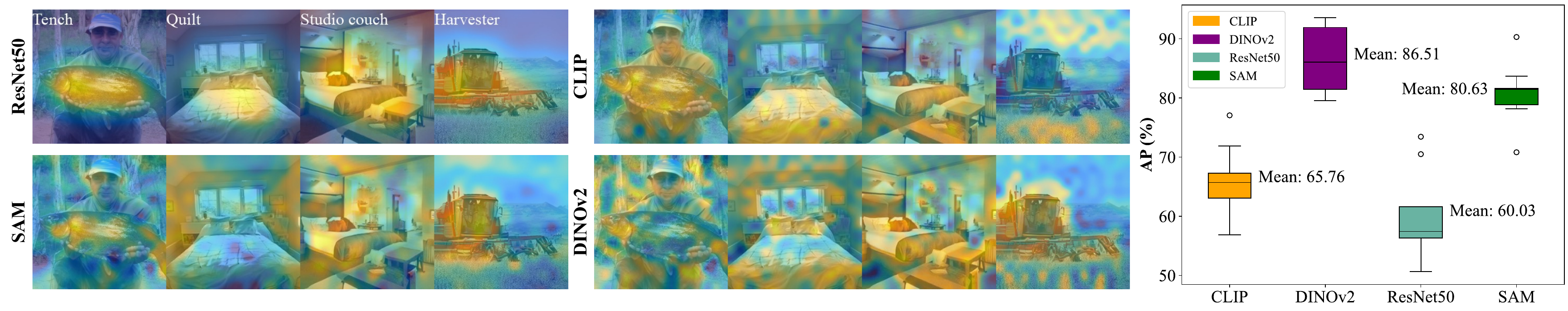}
    \caption{\textbf{Detection performance using different backbones.} The heatmap on the left visualizes what the Fréchet Distance~\cite{fd} perceives for each backbone. The right part shows the detection performance using different backbones.}
    \label{fig:backbone}
\end{figure}

\textbf{Backbone.} 
Fig.~\ref{fig:sim_cam} and Fig.~\ref{fig:backbone} provide visual comparisons of the interest regions identified by different backbones in RIGID and their corresponding performance in detecting AI-generated images.
The heatmaps on the left of Fig.~\ref{fig:backbone} reveal distinct patterns in how each backbone perceives image features:
ResNet50 and CLIP exhibit a more localized focus, highlighting specific regions within the images.
SAM~\cite{sam} and DINOv2 show a more balanced focus, capturing both local details and global context.
The boxplot on the right of Fig.~\ref{fig:backbone} compares the Average Precision of each backbone in detecting generated images. 
Notably, SAM and DINOv2 adopt a holistic approach to image understanding, achieving significantly higher AP scores than models focusing on local features (ResNet50 and CLIP).
This observation underscores the importance of a holistic view of backbones for effective AI-generated image detection. This finding provides valuable insights into RIGID's choice of backbone.
To further validate RIGID's effectiveness stems from its ability to identify fake features, we select some samples with poor generation quality that can be easily distinguished as generated images by an average person, and visualize the RIGID-focus area by GradCAM. As shown in Fig.~\ref{fig:sim_cam}, the high-heat area represents the area with high similarity in eq.~\ref{eq:rigid}, while low-heat regions indicate low similarity. The visualization result clearly demonstrates that RIGID pinpoints the areas containing obvious artificial features.

\section{Discussion}
\label{sec:discuss}

\textbf{Limitations of training-based methods:} While training-based AI-generated image detectors~\cite{corvi, wang, gragnaniello, dire} can perform well under certain conditions, they suffer from the following limitations: (a) \textbf{Expensive training cost.} Training effective detectors demands substantial computational resources and data collection. (b) \textbf{Dependence on quantity and quality of training data.} It can be found in Table~\ref{tb:imagenet} and~\ref{tb:lsun} that detectors with more training samples have higher average performance. However, acquiring a vast collection of high-quality generated images is similarly an expensive task. (c) \textbf{Hyperparameter.} Optimizing training-based detectors requires fine-tuning numerous hyperparameters, such as augmentation methods and related parameters, during training. This process further increases the already substantial training costs. (d) \textbf{Poor generalization.} Table~\ref{tb:imagenet} and~\ref{tb:lsun} clearly show that the training-based detector generalizes poorly to generation styles different from the training data. 

\textbf{Limitations of training-free methods:} Although training-free methods facilitate the problems of high training cost and poor generalization, they also have some limitations. (a) \textbf{Reliance on pretrained models.} Training-free detectors, due to the reliance on pre-trained models, may inherit and perpetuate biases in the original models. For example, AEROBLADE's reliance on LDM autoencoders makes it less effective at detecting images generated using different styles. (b) \textbf{Performance degradation on high-quality generated images.} As shown in Table 1, training-free methods struggle to achieve high detection accuracy on high-quality generated images (e.g., DiT-XL2), although training-based methods perform even worse.
\section{Conclusion}
\label{sec:conclusion}

This paper introduced RIGID, a novel training-free and model-agnostic method for robust detection of AI-generated images. Based on our key observation that real images exhibit less sensitivity to random perturbations in the representation space, RIGID effectively uses this property to distinguish between real and AI-generated images by comparing the representation similarity before and after noise perturbation. Our extensive evaluations demonstrate that RIGID not only surpasses existing training-based and training-free detectors in performance, but also exhibits exceptional generalization across diverse generation methods and resilience to various image corruptions. In terms of \textbf{broader impact}, this research contributes a practical and robust solution to AI-generated image detection, addressing the growing concerns surrounding the potential misuse and harm of GenAI technology.

\newpage
{\small
\bibliographystyle{unsrt}
\bibliography{main}

\begin{thebibliography}{10}

\bibitem{paul2022vision}
Sayak Paul and Pin-Yu Chen.
\newblock Vision transformers are robust learners.
\newblock In {\em Proceedings of the AAAI conference on Artificial Intelligence}, volume~36, pages 2071--2081, 2022.

\bibitem{chin2023prompting4debugging}
Zhi-Yi Chin, Chieh-Ming Jiang, Ching-Chun Huang, Pin-Yu Chen, and Wei-Chen Chiu.
\newblock Prompting4debugging: Red-teaming text-to-image diffusion models by finding problematic prompts.
\newblock {\em International Conference on Machine Learning}, 2024.

\bibitem{genimage}
Mingjian Zhu, Hanting Chen, Qiangyu Yan, Xudong Huang, Guanyu Lin, Wei Li, Zhijun Tu, Hailin Hu, Jie Hu, and Yunhe Wang.
\newblock Genimage: A million-scale benchmark for detecting ai-generated image.
\newblock {\em Advances in Neural Information Processing Systems}, 36, 2024.

\bibitem{wukong}
Jiaxi Gu, Xiaojun Meng, Guansong Lu, Lu~Hou, Niu Minzhe, Xiaodan Liang, Lewei Yao, Runhui Huang, Wei Zhang, Xin Jiang, et~al.
\newblock Wukong: A 100 million large-scale chinese cross-modal pre-training benchmark.
\newblock {\em Advances in Neural Information Processing Systems}, 35:26418--26431, 2022.

\bibitem{sd}
Robin Rombach, Andreas Blattmann, Dominik Lorenz, Patrick Esser, and Bj{\"o}rn Ommer.
\newblock High-resolution image synthesis with latent diffusion models.
\newblock In {\em Proceedings of the IEEE/CVF conference on computer vision and pattern recognition}, pages 10684--10695, 2022.

\bibitem{leaderboard}
Papers with Code.
\newblock https://paperswithcode.com/task/image-generation.

\bibitem{midjourney}
Midjourney.
\newblock https://www.midjourney.com/home/.
\newblock 2022.

\bibitem{exposing}
George Stein, Jesse Cresswell, Rasa Hosseinzadeh, Yi~Sui, Brendan Ross, Valentin Villecroze, Zhaoyan Liu, Anthony~L Caterini, Eric Taylor, and Gabriel Loaiza-Ganem.
\newblock Exposing flaws of generative model evaluation metrics and their unfair treatment of diffusion models.
\newblock {\em Advances in Neural Information Processing Systems}, 36, 2024.

\bibitem{imagenet}
Jia Deng, Wei Dong, Richard Socher, Li-Jia Li, Kai Li, and Li~Fei-Fei.
\newblock Imagenet: A large-scale hierarchical image database.
\newblock In {\em 2009 IEEE conference on computer vision and pattern recognition}, pages 248--255. Ieee, 2009.

\bibitem{dinov2}
Maxime Oquab, Timoth{\'e}e Darcet, Th{\'e}o Moutakanni, Huy Vo, Marc Szafraniec, Vasil Khalidov, Pierre Fernandez, Daniel Haziza, Francisco Massa, Alaaeldin El-Nouby, et~al.
\newblock Dinov2: Learning robust visual features without supervision.
\newblock {\em arXiv preprint arXiv:2304.07193}, 2023.

\bibitem{lsun}
Fisher Yu, Ari Seff, Yinda Zhang, Shuran Song, Thomas Funkhouser, and Jianxiong Xiao.
\newblock Lsun: Construction of a large-scale image dataset using deep learning with humans in the loop.
\newblock {\em arXiv preprint arXiv:1506.03365}, 2015.

\bibitem{clip}
Alec Radford, Jong~Wook Kim, Chris Hallacy, Aditya Ramesh, Gabriel Goh, Sandhini Agarwal, Girish Sastry, Amanda Askell, Pamela Mishkin, Jack Clark, et~al.
\newblock Learning transferable visual models from natural language supervision.
\newblock In {\em International conference on machine learning}, pages 8748--8763. PMLR, 2021.

\bibitem{vit}
Alexey Dosovitskiy, Lucas Beyer, Alexander Kolesnikov, Dirk Weissenborn, Xiaohua Zhai, Thomas Unterthiner, Mostafa Dehghani, Matthias Minderer, Georg Heigold, Sylvain Gelly, et~al.
\newblock An image is worth 16x16 words: Transformers for image recognition at scale.
\newblock {\em arXiv preprint arXiv:2010.11929}, 2020.

\bibitem{dit}
William Peebles and Saining Xie.
\newblock Scalable diffusion models with transformers.
\newblock In {\em Proceedings of the IEEE/CVF International Conference on Computer Vision}, pages 4195--4205, 2023.

\bibitem{adm}
Prafulla Dhariwal and Alexander Nichol.
\newblock Diffusion models beat gans on image synthesis.
\newblock {\em Advances in neural information processing systems}, 34:8780--8794, 2021.

\bibitem{biggan}
Andrew Brock, Jeff Donahue, and Karen Simonyan.
\newblock Large scale gan training for high fidelity natural image synthesis.
\newblock {\em arXiv preprint arXiv:1809.11096}, 2018.

\bibitem{gigagan}
Minguk Kang, Jun-Yan Zhu, Richard Zhang, Jaesik Park, Eli Shechtman, Sylvain Paris, and Taesung Park.
\newblock Scaling up gans for text-to-image synthesis.
\newblock In {\em Proceedings of the IEEE/CVF Conference on Computer Vision and Pattern Recognition}, pages 10124--10134, 2023.

\bibitem{ldm}
Robin Rombach, Andreas Blattmann, Dominik Lorenz, Patrick Esser, and Bj{\"o}rn Ommer.
\newblock High-resolution image synthesis with latent diffusion models.
\newblock In {\em Proceedings of the IEEE/CVF conference on computer vision and pattern recognition}, pages 10684--10695, 2022.

\bibitem{maskgit}
Huiwen Chang, Han Zhang, Lu~Jiang, Ce~Liu, and William~T Freeman.
\newblock Maskgit: Masked generative image transformer.
\newblock In {\em Proceedings of the IEEE/CVF Conference on Computer Vision and Pattern Recognition}, pages 11315--11325, 2022.

\bibitem{rqtransformer}
Doyup Lee, Chiheon Kim, Saehoon Kim, Minsu Cho, and Wook-Shin Han.
\newblock Autoregressive image generation using residual quantization.
\newblock In {\em Proceedings of the IEEE/CVF Conference on Computer Vision and Pattern Recognition}, pages 11523--11532, 2022.

\bibitem{styleganxl}
Axel Sauer, Katja Schwarz, and Andreas Geiger.
\newblock Stylegan-xl: Scaling stylegan to large diverse datasets.
\newblock In {\em ACM SIGGRAPH 2022 conference proceedings}, pages 1--10, 2022.

\bibitem{ddpm}
Jonathan Ho, Ajay Jain, and Pieter Abbeel.
\newblock Denoising diffusion probabilistic models.
\newblock {\em Advances in neural information processing systems}, 33:6840--6851, 2020.

\bibitem{iddpm}
Alexander~Quinn Nichol and Prafulla Dhariwal.
\newblock Improved denoising diffusion probabilistic models.
\newblock In {\em International conference on machine learning}, pages 8162--8171. PMLR, 2021.

\bibitem{stylegan}
Tero Karras, Samuli Laine, and Timo Aila.
\newblock A style-based generator architecture for generative adversarial networks.
\newblock In {\em Proceedings of the IEEE/CVF conference on computer vision and pattern recognition}, pages 4401--4410, 2019.

\bibitem{unleashing}
Sam Bond-Taylor, Peter Hessey, Hiroshi Sasaki, Toby~P Breckon, and Chris~G Willcocks.
\newblock Unleashing transformers: Parallel token prediction with discrete absorbing diffusion for fast high-resolution image generation from vector-quantized codes.
\newblock In {\em European Conference on Computer Vision}, pages 170--188. Springer, 2022.

\bibitem{projdiffusiongan}
Zhendong Wang, Huangjie Zheng, Pengcheng He, Weizhu Chen, and Mingyuan Zhou.
\newblock Diffusion-gan: Training gans with diffusion.
\newblock {\em arXiv preprint arXiv:2206.02262}, 2022.

\bibitem{projgan}
Axel Sauer, Kashyap Chitta, Jens M{\"u}ller, and Andreas Geiger.
\newblock Projected gans converge faster.
\newblock {\em Advances in Neural Information Processing Systems}, 34:17480--17492, 2021.

\bibitem{corvi}
Riccardo Corvi, Davide Cozzolino, Giada Zingarini, Giovanni Poggi, Koki Nagano, and Luisa Verdoliva.
\newblock On the detection of synthetic images generated by diffusion models.
\newblock In {\em ICASSP 2023-2023 IEEE International Conference on Acoustics, Speech and Signal Processing (ICASSP)}, pages 1--5. IEEE, 2023.

\bibitem{gragnaniello}
Diego Gragnaniello, Davide Cozzolino, Francesco Marra, Giovanni Poggi, and Luisa Verdoliva.
\newblock Are gan generated images easy to detect? a critical analysis of the state-of-the-art.
\newblock In {\em 2021 IEEE international conference on multimedia and expo (ICME)}, pages 1--6. IEEE, 2021.

\bibitem{dire}
Zhendong Wang, Jianmin Bao, Wengang Zhou, Weilun Wang, Hezhen Hu, Hong Chen, and Houqiang Li.
\newblock Dire for diffusion-generated image detection.
\newblock In {\em Proceedings of the IEEE/CVF International Conference on Computer Vision}, pages 22445--22455, 2023.

\bibitem{wang}
Sheng-Yu Wang, Oliver Wang, Richard Zhang, Andrew Owens, and Alexei~A Efros.
\newblock Cnn-generated images are surprisingly easy to spot... for now.
\newblock In {\em Proceedings of the IEEE/CVF conference on computer vision and pattern recognition}, pages 8695--8704, 2020.

\bibitem{aeroblade}
Jonas Ricker, Denis Lukovnikov, and Asja Fischer.
\newblock Aeroblade: Training-free detection of latent diffusion images using autoencoder reconstruction error.
\newblock In {\em Proceedings of the {IEEE} Conference on Computer Vision and Pattern Recognition ({CVPR})}, 2024.

\bibitem{lpips}
Richard Zhang, Phillip Isola, Alexei~A Efros, Eli Shechtman, and Oliver Wang.
\newblock The unreasonable effectiveness of deep features as a perceptual metric.
\newblock In {\em Proceedings of the IEEE conference on computer vision and pattern recognition}, pages 586--595, 2018.

\bibitem{patch}
Nan Zhong, Yiran Xu, Zhenxing Qian, and Xinpeng Zhang.
\newblock Rich and poor texture contrast: A simple yet effective approach for ai-generated image detection.
\newblock {\em arXiv preprint arXiv:2311.12397}, 2023.

\bibitem{jailbreak1}
Yu-Lin Tsai, Chia-Yi Hsu, Chulin Xie, Chih-Hsun Lin, Jia~You Chen, Bo~Li, Pin-Yu Chen, Chia-Mu Yu, and Chun-Ying Huang.
\newblock Ring-a-bell! how reliable are concept removal methods for diffusion models?
\newblock In {\em The Twelfth International Conference on Learning Representations}, 2023.

\bibitem{jailbreak2}
Yijun Yang, Ruiyuan Gao, Xiaosen Wang, Tsung-Yi Ho, Nan Xu, and Qiang Xu.
\newblock {MMA-Diffusion: MultiModal Attack on Diffusion Models}.
\newblock In {\em Proceedings of the {IEEE} Conference on Computer Vision and Pattern Recognition ({CVPR})}, 2024.

\bibitem{artifact1}
Joel Frank, Thorsten Eisenhofer, Lea Sch{\"o}nherr, Asja Fischer, Dorothea Kolossa, and Thorsten Holz.
\newblock Leveraging frequency analysis for deep fake image recognition.
\newblock In {\em International conference on machine learning}, pages 3247--3258. PMLR, 2020.

\bibitem{artifact2}
Tarik Dzanic, Karan Shah, and Freddie Witherden.
\newblock Fourier spectrum discrepancies in deep network generated images.
\newblock {\em Advances in neural information processing systems}, 33:3022--3032, 2020.

\bibitem{artifact3}
Keshigeyan Chandrasegaran, Ngoc-Trung Tran, and Ngai-Man Cheung.
\newblock A closer look at fourier spectrum discrepancies for cnn-generated images detection.
\newblock In {\em Proceedings of the IEEE/CVF conference on computer vision and pattern recognition}, pages 7200--7209, 2021.

\bibitem{ojha}
Utkarsh Ojha, Yuheng Li, and Yong~Jae Lee.
\newblock Towards universal fake image detectors that generalize across generative models.
\newblock In {\em Proceedings of the IEEE/CVF Conference on Computer Vision and Pattern Recognition}, pages 24480--24489, 2023.

\bibitem{landscape}
Hao Li, Zheng Xu, Gavin Taylor, Christoph Studer, and Tom Goldstein.
\newblock Visualizing the loss landscape of neural nets.
\newblock {\em Advances in neural information processing systems}, 31, 2018.

\bibitem{gradcam}
Ramprasaath~R Selvaraju, Michael Cogswell, Abhishek Das, Ramakrishna Vedantam, Devi Parikh, and Dhruv Batra.
\newblock Grad-cam: Visual explanations from deep networks via gradient-based localization.
\newblock In {\em Proceedings of the IEEE international conference on computer vision}, pages 618--626, 2017.

\bibitem{gan}
Ian Goodfellow, Jean Pouget-Abadie, Mehdi Mirza, Bing Xu, David Warde-Farley, Sherjil Ozair, Aaron Courville, and Yoshua Bengio.
\newblock Generative adversarial networks.
\newblock {\em Communications of the ACM}, 63(11):139--144, 2020.

\bibitem{color}
Scott McCloskey and Michael Albright.
\newblock Detecting gan-generated imagery using color cues.
\newblock {\em arXiv preprint arXiv:1812.08247}, 2018.

\bibitem{saturation}
Scott McCloskey and Michael Albright.
\newblock Detecting gan-generated imagery using saturation cues.
\newblock In {\em 2019 IEEE international conference on image processing (ICIP)}, pages 4584--4588. IEEE, 2019.

\bibitem{cooccurrence}
Lakshmanan Nataraj, Tajuddin~Manhar Mohammed, Shivkumar Chandrasekaran, Arjuna Flenner, Jawadul~H Bappy, Amit~K Roy-Chowdhury, and BS~Manjunath.
\newblock Detecting gan generated fake images using co-occurrence matrices.
\newblock {\em arXiv preprint arXiv:1903.06836}, 2019.

\bibitem{ddim}
Jiaming Song, Chenlin Meng, and Stefano Ermon.
\newblock Denoising diffusion implicit models.
\newblock In {\em International Conference on Learning Representations}, 2020.

\bibitem{progan}
Tero Karras, Timo Aila, Samuli Laine, and Jaakko Lehtinen.
\newblock Progressive growing of gans for improved quality, stability, and variation.
\newblock In {\em International Conference on Learning Representations}, 2018.

\bibitem{stein}
Charles~M Stein.
\newblock Estimation of the mean of a multivariate normal distribution.
\newblock {\em The annals of Statistics}, pages 1135--1151, 1981.

\bibitem{sam}
Alexander Kirillov, Eric Mintun, Nikhila Ravi, Hanzi Mao, Chloe Rolland, Laura Gustafson, Tete Xiao, Spencer Whitehead, Alexander~C Berg, Wan-Yen Lo, et~al.
\newblock Segment anything.
\newblock In {\em Proceedings of the IEEE/CVF International Conference on Computer Vision}, pages 4015--4026, 2023.

\bibitem{rs}
Jeremy Cohen, Elan Rosenfeld, and Zico Kolter.
\newblock Certified adversarial robustness via randomized smoothing.
\newblock In {\em international conference on machine learning}, pages 1310--1320. PMLR, 2019.

\bibitem{fd}
Martin Heusel, Hubert Ramsauer, Thomas Unterthiner, Bernhard Nessler, and Sepp Hochreiter.
\newblock Gans trained by a two time-scale update rule converge to a local nash equilibrium.
\newblock {\em Advances in neural information processing systems}, 30, 2017.

\bibitem{resnet}
Kaiming He, Xiangyu Zhang, Shaoqing Ren, and Jian Sun.
\newblock Deep residual learning for image recognition.
\newblock In {\em Proceedings of the IEEE conference on computer vision and pattern recognition}, pages 770--778, 2016.

\end{thebibliography}
}

\newpage
\appendix

\section{Experimental Details}
\label{ap:setting}
All our experiments were tested on a NVIDIA GeForce RTX 3090 with 24G memory. The model we used is DINOv2~\cite{dinov2} VIT Large with a patch size of 14, and the noise intensity $\lambda$ is 0.05.

\section{Cosine Similarity Landscape}
\label{ap:landscape}

Following~\cite{landscape}, we plot the cosine similarity landscape of real and generated images. The plot function is defined as follows:

\begin{equation}
    f(x | \alpha, \beta) = \frac{1}{|X|} \sum_{x \in X} \textsf{sim}[f_{\theta}(x \oplus (\alpha \textbf{u} + \beta \textbf{v})), f_{\theta}(x)]
    \label{eq:landscape}
\end{equation}

Where $X$ represents the sample set of real images or generated images, $sim$ is the cosine similarity, $f_{\theta}(\cdot)$ is a feature extractor, and $\textbf{u}$ and $\textbf{v}$ are two random direction vectors sampled from the Gaussian distribution. We plot the cosine similarity landscape of ResNet50, CLIP and DINOv2 in Fig.~\ref{fig:framework}. In our experiments, $\alpha$ and $\beta$ range from -0.5 to 0.5 with a step size of 0.01.

\begin{figure*}[t]
    \includegraphics[width=\linewidth]{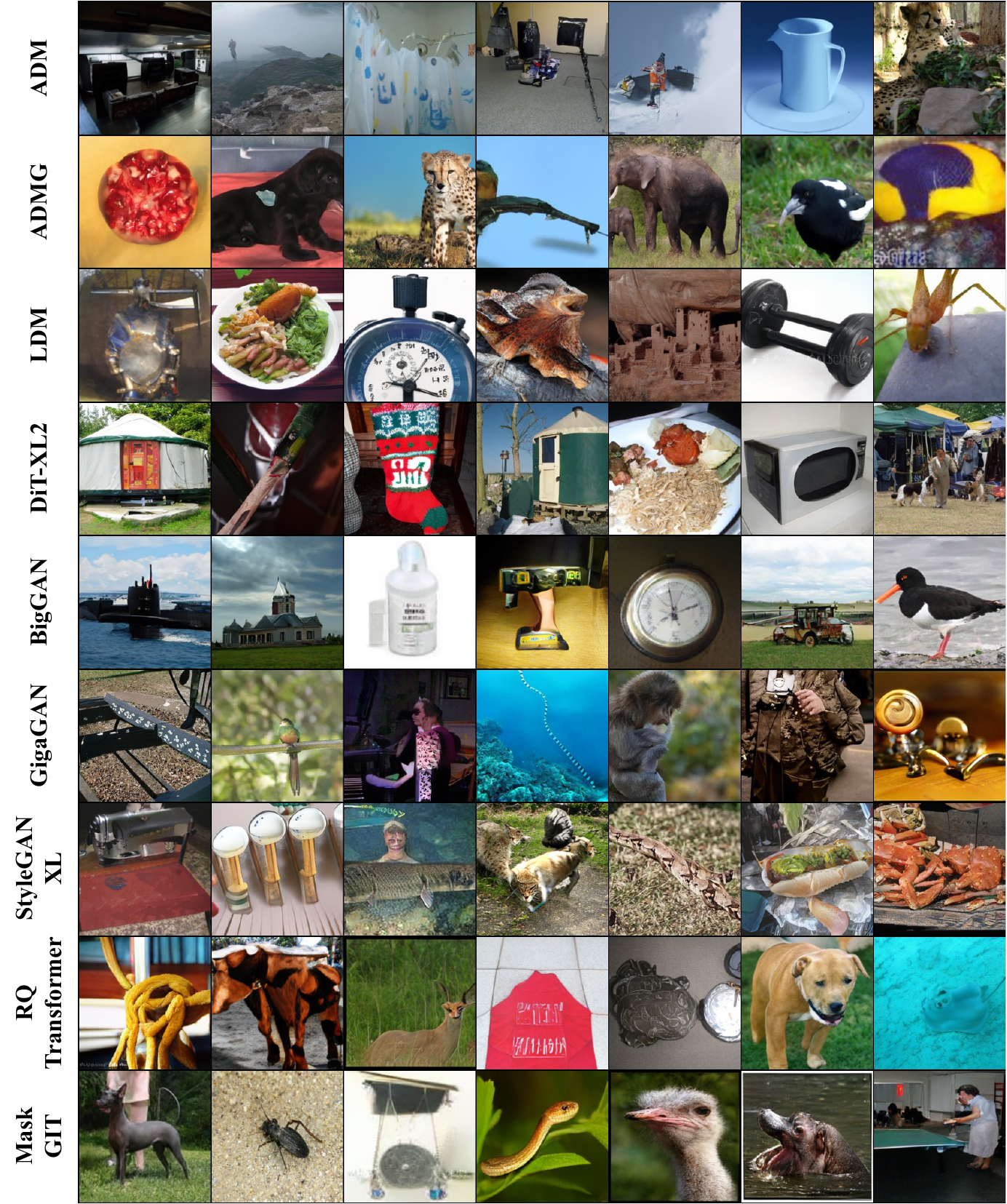}
\caption{\textbf{Display of Generated Images on \imagenet.} Generation methods include: ADM, ADMG, LDM, DiT-XL2, BigGAN, GigaGAN, StyleGAN-XL, RQ-Transformer and MaskGIT. }
\label{fig:gen_imagenet}
\end{figure*}

\begin{figure*}[t]
    \includegraphics[width=\linewidth]{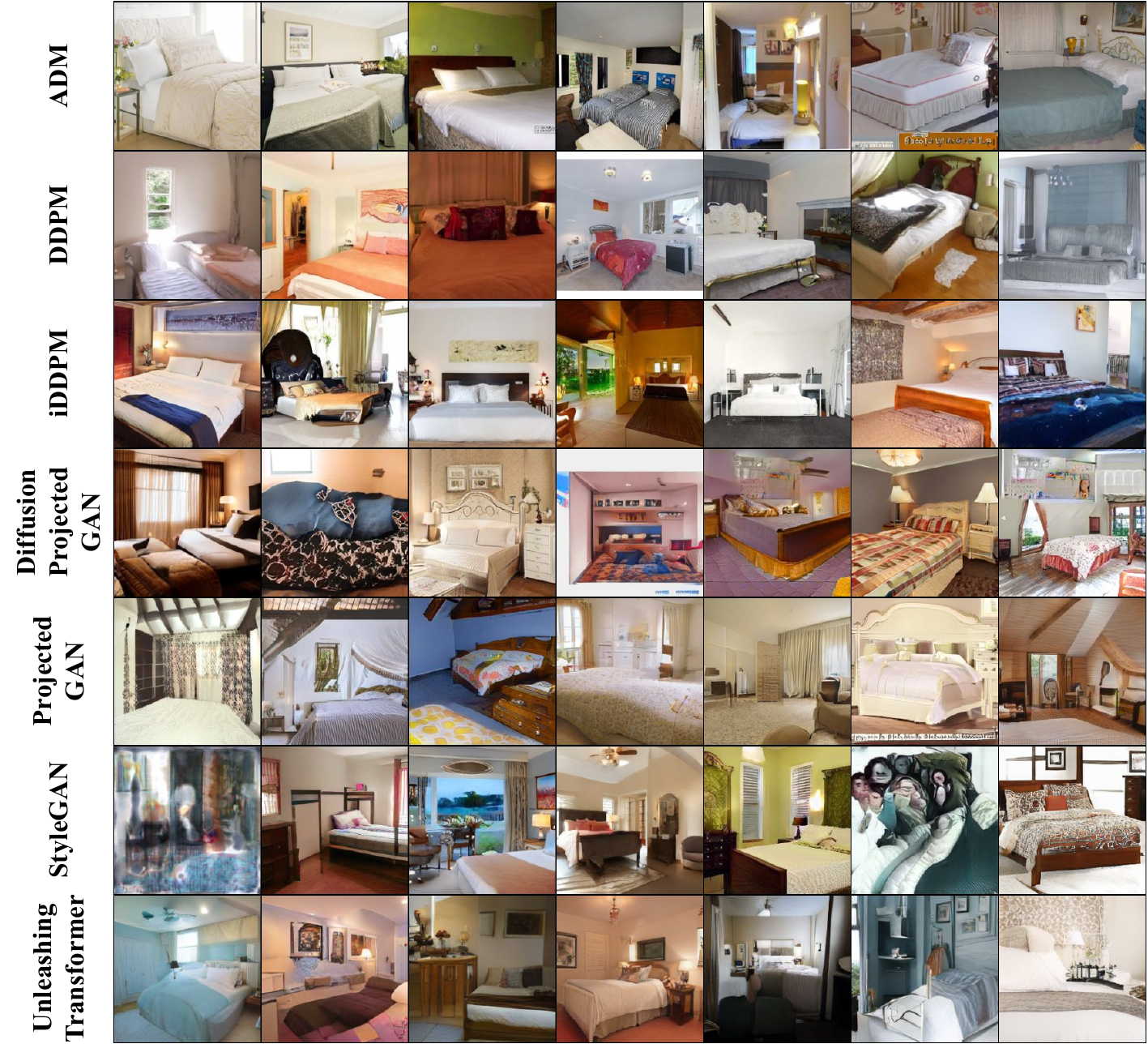}
\caption{\textbf{Display of Generated Images on \lsun.} Generation methods include: ADM, DDPM, iDDPM, Diffusion Projected GAN, Projected GAN, StyleGAN and Unleashing Transformer.}
\label{fig:gen_lsun}
\end{figure*}


\section{Generated Datasets}
\label{ap:datasets}

The generated images on \imagenet and \lsun we used are both from~\cite{exposing}, which generated 100,000 images for each generation model in each dataset based on the leaderboard~\cite{leaderboard} of generation quality on the two datasets. For class-conditional models, the same number of samples from each class is generated, i.e. 100 images per class in \imagenet.
The repository link and FID scores of different generation methods on \imagenet and \lsun are as follows:

\subsection{\imagenet}

\begin{itemize}
    \item Three models used sets of 50k publicly available images provided at \url{https://github.com/ openai/guided-diffusion/tree/main/evaluations}
    \begin{itemize}
        \item \textbf{ADM}~\cite{adm}. FID=11.84
        \item \textbf{ADMG}~\cite{adm}. FID=5.58
        \item \textbf{BigGAN}~\cite{biggan}. FID=7.94
    \end{itemize}
    \item \textbf{DiT-XL-2}~\cite{dit}. FID=2.80. \url{https://github.com/facebookresearch/DiT}.
    \item \textbf{GigaGAN}~\cite{gigagan}. With 100k images provided privately by authors. FID=4.16.
    \item \textbf{LDM}~\cite{ldm}. FID=4.29. \url{https://github.com/CompVis/latent-diffusion}.
    \item \textbf{StyleGAN-XL}~\cite{styleganxl}. FID=2.91. \url{https://github.com/autonomousvision/stylegan-xl}.
    \item \textbf{RQ-Transformer}~\cite{rqtransformer}. FID=9.71. \url{https://github.com/kakaobrain/rq-vae-transformer}.
    \item \textbf{Mask-GIT}~\cite{maskgit}. FID=5.63. \url{https://github.com/google-research/maskgit}.
\end{itemize}

\subsection{\lsun}

\begin{itemize}
    \item Three models used sets of 50k publicly available images provided at \url{https://github.com/ openai/guided-diffusion/tree/main/evaluations}.
    \begin{itemize}
        \item \textbf{ADM}~\cite{adm}. FID=2.20
        \item \textbf{DDPM}~\cite{ddpm}. FID=5.18.
        \item \textbf{iDDPM}~\cite{iddpm}. FID=4.54.
        \item \textbf{StyleGAN}~\cite{stylegan}. FID=2.65.
    \end{itemize}
    \item \textbf{Diffusion-Projected GAN}~\cite{projdiffusiongan}. FID=1.79. \url{https://github.com/Zhendong-Wang/Diffusion-GAN}. 
    \item \textbf{Projected GAN}~\cite{projgan}. FID=2.23. \url{https://github.com/autonomousvision/projected-gan}. 
    \item \textbf{Unleashing Transformers}~\cite{unleashing}. FID=3.58. \url{https://github.com/samb-t/unleashing-transformers}. 
\end{itemize}

\subsection{GenImage}
GenImage~\cite{genimage} is the latest million-level benchmark for detecting AI-generated images. One of the advantages of GenImage is that it contains generated images from four mainstream text-to-image platforms, including: Wukong~\cite{wukong}, SD 1.4~\cite{sd}, SD 1.5~\cite{sd} and Midjourney~\cite{midjourney}.
GenImage input sentences follow the template "photo of class", where "class" is replaced by ImageNet labels. For Wukong, Chinese sentences tend to achieve better generation quality. In this way, the sentences are translated into Chinese in advance.

\section{Baselines}
\label{ap:baseline}

\textbf{Wang et al.}~\cite{wang} We use the code and model checkpoints from the official repository\footnote{https://github.com/PeterWang512/CNNDetection}.

\textbf{Gragnaniello et al.}~\cite{gragnaniello} and \textbf{Corvi et al.}~\cite{corvi} we use the code and model checkpoints from the official repository\footnote{https://github.com/grip-unina/DMimageDetection} provided by Corvi et al. This repository also includes the detector from Gragnaniello et al.

\textbf{DIRE}~\cite{dire} We use the code and model checkpoints from the official repository\footnote{https://github.com/ZhendongWang6/DIRE}. However, \cite{aeroblade} points out that the excellent performance reported in DIRE is because it saves real images as jpegs and generated images as png, which causes DIRE to learn the differences between formats. Therefore, we converted both real images and generated images into jpeg format and tested their performance as shown in Tables~\ref{tb:imagenet} and~\ref{tb:lsun}.

\textbf{AEROBLADE}~\cite{aeroblade} We use the code from the official repository\footnote{https://github.com/jonasricker/aeroblade}. We use the autoencoder from CompVis-stable-diffusion-v1-1-ViT-L-14-openai to compute the reconstruction error.

\section{Display of Generated Images}
We display images generated by different generation methods on \imagenet and \lsun in Fig.~\ref{fig:gen_imagenet} and Fig.~\ref{fig:gen_lsun}.

\section{Display of Perturbed Images}
\begin{figure*}[t]
    \includegraphics[width=\linewidth]{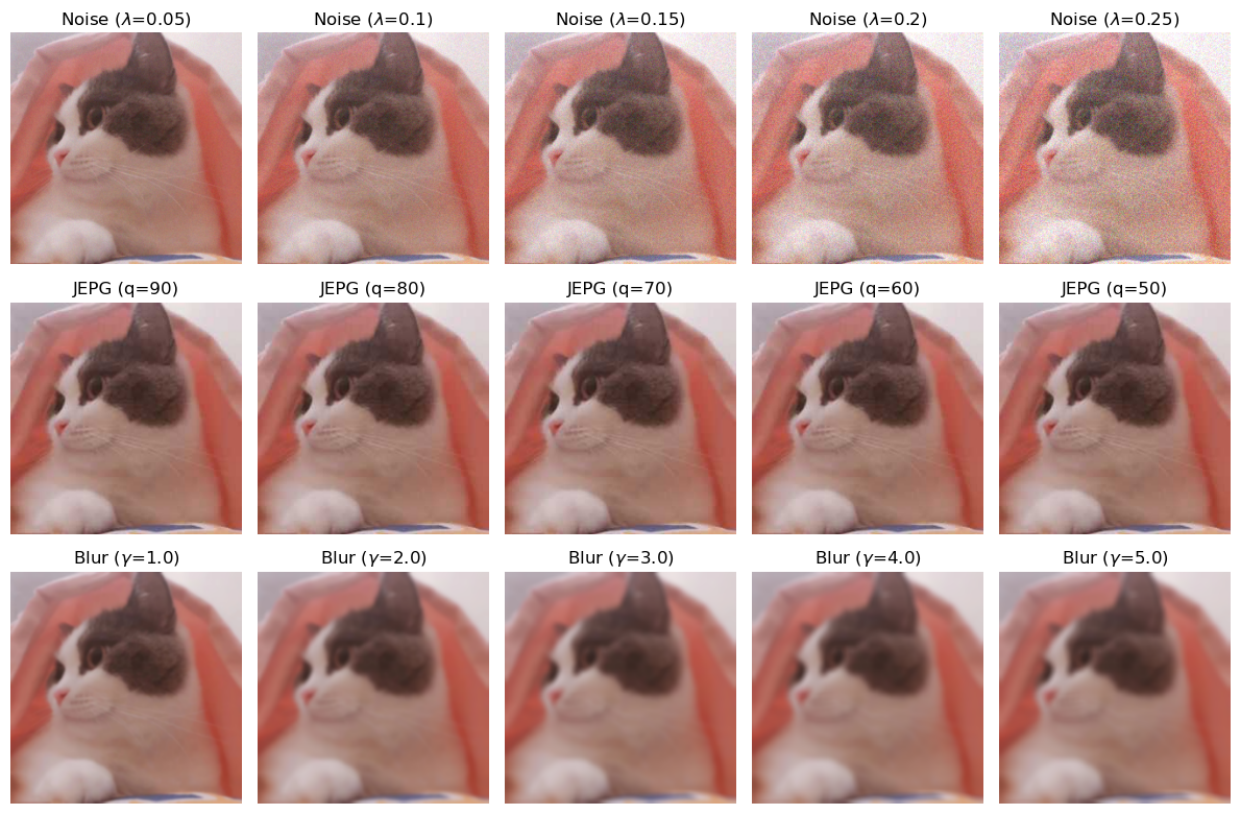}
\caption{\textbf{Display of Perturbed Images.} The first row shows the images perturbed by Gaussian noise with different intensities $\lambda$. The second row shows the JPEG compressed images with various qualities and the bottom row shows the Gaussian blurred images.}
\label{fig:display_perturb}
\end{figure*}

We display images perturbed by different 3 perturbation methods: Gaussian Noise, JPEG Compression and Gaussian Blur in Fig.~\ref{fig:display_perturb}. For each perturbation, we set five levels, including $\lambda = 0.05, 0.1, 0.15, 0.2, 0.25$, $q = 90, 80, 70, 60, 50$ and $\gamma = 1.0, 2.0, 3.0, 4.0, 5.0$.

\section{Ablation Study: Noise}

In Sec.~\ref{sec:ablation}, we discuss the impact of perturbation intensity and backbone model on \rigid detection performance. Further, we compare the impact of noise from different distributions on the performance of \rigid in Table~\ref{tb:noise}. The distributions we use include: Laplace distribution, Gamma distribution, Chi-square distribution and Gaussian distribution. We fix the noise intensity to 0.05. It can be seen that using different noises has a minimal impact on the overall performance of \rigid. 

\begin{table}[t]
\caption{The AP of noise from different distribution on \imagenet. A higher value indicates better performance.}
\setlength\tabcolsep{2pt}
\renewcommand{\arraystretch}{1.3}
\resizebox{1.\linewidth}{!}{
\begin{tabular}{ccccccccccc}
\hline
\textbf{Distribution} & \textbf{ADM} & \textbf{ADMG} & \textbf{LDM} & \textbf{DiT} & \textbf{BigGAN} & \textbf{GigaGAN} & \textbf{StyleGAN XL} & \textbf{RQ-Transformer} & \textbf{Mask GIT} & \textbf{Aver} \\ \hline
Laplace               & 86.36        & 79.49         & 78.57        & 67.91        & 93.98           & 86.49            & 84.53                & 92.65                   & 90.94             & 84.55         \\
Gamma                 & 85.96        & 80.51         & 78.58        & 71.82        & 93.15           & 88.70            & 84.73                & 93.24                   & 90.82             & 85.28         \\
Chi-Square            & 86.65        & 79.74         & 75.86        & 68.09        & 94.76           & 88.25            & 86.42                & 92.73                   & 91.45             & 84.88         \\
Gaussian              & 86.06        & 81.46         & 80.23        & 69.55        & 93.57           & 87.92            & 84.75                & 93.11                   & 91.91             & 85.40         \\ \hline
\end{tabular}}
\label{tb:noise}
\end{table}

\end{document}